\documentclass{article}
\usepackage{spconf,amsmath,graphicx}

\usepackage{todonotes}
\usepackage{hyperref}
\usepackage[T1]{fontenc}

\usepackage{caption}
\usepackage{subcaption}

\title{Make More of Your Data: Minimal Effort Data Augmentation for Automatic Speech Recognition and Translation} % Effortless?
\name{Tsz Kin Lam$^{\star}$ \qquad Shigehiko Schamoni$^{\star, \dagger}$ \qquad Stefan Riezler$^{\star,\dagger}$}
\address{$^{\star}$Computational Linguistics \& $^\dagger$IWR, Heidelberg University, Germany \\
  {\tt \{lam,schamoni,riezler\}@cl.uni-heidelberg.de}
}

\begin{document}
%\ninept
%
\maketitle
%
%{\footnotesize
%
\begin{abstract}
Data augmentation is a technique to generate new training data based on existing data. We evaluate the simple and cost-effective method of concatenating the original data examples to build new training instances. Continued training with such augmented data is able to improve off-the-shelf Transformer and Conformer models that were optimized on the original data only. We demonstrate considerable improvements on the LibriSpeech-960h test sets (WER 2.83 and 6.87 for \texttt{test-clean} and \texttt{test-other}), which carry over to models combined with shallow fusion (WER 2.55 and 6.27). Our method of continued training also leads to improvements of up to 0.9 WER on the ASR part of CoVoST-2 for four non-English languages, and we observe that the gains are highly dependent on the size of the original training data. We compare different concatenation strategies and found that our method does not need speaker information to achieve its improvements. Finally, we demonstrate on two datasets that our methods also works for speech translation tasks. 

%A final ablation experiment sheds some light on the sensitivity of these models towards ill-posed\todo{= property of equations in sense of Hadamard, relation to data?} data examples.

%This paper proposes \texttt{FineCat}, a simple but effective concatenation based data augmentation method for fine-tuning attention based speech-to-text models. \texttt{FineCat} creates effective training data by combining training instances in temporal dimension. We experiment costless concatenation strategies such as 1) \texttt{Random}, 2) \texttt{Speaker-based} and 3) \texttt{Copying} in speech-to-text tasks. Our experiments show that \texttt{Random} and \texttt{Speaker-based} obtain similar improvement over the baseline but \texttt{Copying} enhance repetition behaviour. \texttt{FineCat} is complementary to shallow fusion as demonstrated by experiments in LibriSpeech-960h data. It also improves non-English ASR models as demonstrated by experiments in CoVoST-2 ASR data. We also examine its usage on AST on MuST-C and CoVoST-2 En-De language direction.
\end{abstract}
% ----------------------------------------------------------
% ----------------------------------------------------------
%\vspace{-1mm}
\section{Introduction}
%\vspace{-2mm}
\label{sec:intro}
% ----------------------------------------------------------
%\todo{TK: Possibly merge and condense information in introduction+relatedWork. They take too much space}
Data augmentation (DA) is an active research field in many machine learning areas. 
%It has to solve the problem of balancing the requirement for large and informative datasets to train complex neural models with the requirement of avoiding exploding training costs and overfitting.
%It tackles the problem of balancing the requirement for large and informative datasets with the huge efforts and escalating costs to generate such datasets. 
%The rapid development of machine learning in recent years and the availability of complex neural networks not only intensifies this requirement but also exacerbates problems such as overfitting and exploding training costs.
It addresses the problem of creating large and informative datasets for data-hungry neural networks with automatic techniques. 
The standard technique for DA is extending or enhancing training data, such that the final model quality is improved either by simply increasing the total amount of training data, or by creating informative training instances that improve robustness. In practice, both aspects are interwoven and work together. 

The situation in speech-to-text processing such as automatic speech recognition (ASR) is particularly difficult due to two main reasons: First, generating speech data sets is expensive, and second, training of ASR models is extremely resource intensive. Thus, it is desirable to make most of the data and of pre-trained models that are available.%, and to improve on off-she-shelf models that were trained and published by large tech-companies that have the resources to optimize their own models to perfection (CITATION).

We evaluate the applicability of one of simplest DA techniques, namely concatenating training instances of the original data to create new training instances, to speech-to-text processing. Our method does not need any additional data or resources, and comes with low computational effort that allows applying the augmentation procedure {in-memory} and {on-the-fly}. 
Our experiments show that already very strong models %that were trained and optimized by large tech-companies 
can be further improved with {continued training} using a concatenation based DA approach. We further evaluate different strategies for selecting data to concatenate, and find that these strategies can make a difference depending on the size and complexity of the data set. Furthermore, we show that it is important to combine augmented data with the original to prevent degradation during continued training.

Our results are evaluated on the LibriSpeech-960h data, %\todo{mention WER results on LibriSpeech here}, 
with and without shallow fusion \cite{toshniwal2018comparison}, i.e., the integration of an external language model (LM) in the decoding step, 
where our method is able to reduce WER down to 2.55 and 6.27 on \texttt{test-clean} and \texttt{test-other}, respectively.
We also conduct experiments on the ASR part of the CoVoST-2 data set for five languages, namely English, German, Catalan, French and Spanish, and show absolute improvements of up to 0.9 WER points. % and found that the improvements are highly dependent on the amount of original training data, where models that were already trained on large amounts of data benefit less from our augmentation strategies.
%Finally, we found significant improvements when applying our method on automatic speech translation (AST) tasks.

%Call it ICE: Inject, Combine and Eliminate \\
%Inject: when to inject the augmented data, e.g., at the beginning or fine-tuning. 
%Combine: How to combine the data, e.g. by Random, by Speaker or By Copy?
%Eliminate: Just filtering though we filter it based on length only. 
%Tsz Kin: \textbf{fine-tuning} seems to be dangerous because it usually involves domain adaptation. \\
%let's call it C$^{3}$ cost-less continual training by concatenation

\begin{table*}[ht]
\centering
%\resizebox{.48\textwidth}{!}{%
\begin{tabular}{lcccc}
\hline
\textbf{Model} & \texttt{test-clean} & {w/shallow fusion} & \texttt{test-other} & {w/shallow fusion} \\
\hline
Pre-trained & $3.30$ & $3.13$ & $7.51$ & $6.81$ \\
\hline
%CT orig $\cup$ CatSelf & $3.81$ & $4.24$ & $7.97$ & $7.49$ \\
CT orig $\cup$ CatSpeaker & $2.83$ {\footnotesize $\pm 0.03$} & $2.55$ {\footnotesize $\pm 0.04$} & $6.87$ {\footnotesize $\pm 0.03$} & $6.27$ {\footnotesize $\pm 0.07$}\\
CT orig $\cup$ CatRandom & $2.90$ {\footnotesize $\pm 0.01$} & $2.65$ {\footnotesize $\pm 0.02$} & $6.93$ {\footnotesize $\pm  0.06$} & $6.36$ {\footnotesize $\pm 0.09$}\\
\hline
\end{tabular}
%}
\caption{Word Error Rate of \textit{pre-trained} and \textit{continued training (CT)} ASR models on LibriSpeech \texttt{test-clean} and \texttt{test-other} data sets with and without shallow fusion (SF). The ``$\pm$'' values indicate standard deviation over 3 runs.
%Both CT orig $\cup$ CatSpeaker and CT orig $\cup$ CatRandom are repeated with three random seeds.
}
\label{tab:asr_librispeech}
\vspace{-5mm}
\end{table*}

{
\begin{table}[t]
\centering
\resizebox{.48\textwidth}{!}{%

\begin{tabular}{lcccc}
\hline
\textbf{Model} & {\footnotesize\texttt{test-clean}} & {\footnotesize with SF} & {\footnotesize\texttt{test-other}} & {\footnotesize with SF} \\
\hline
Pre-trained & $3.30$ & $3.13$ & $7.51$ & $6.81$ \\
\hline
%CT orig & $3.26$ {\footnotesize $\pm 0.03$} & $3.05$ {\footnotesize $\pm 0.01$} & $7.38$ {\footnotesize $\pm 0.03$} & $6.82$ {\footnotesize $\pm 0.07$} \\
CT orig & $3.26$ & $3.05$ & $7.38$ & $6.82$ \\
%CT CatSelf & $41.29$ & $46.48$ & $54.31$ & $57.80$\\
CT CatSpeaker & $2.94$ & $2.55$ & $7.09$ & $6.42$ \\
CT CatRandom & $2.94$ & $2.64$ & $7.31$ & $6.51$ \\
\hline
\end{tabular}
}
\caption{Ablation experiment: Word Error Rate of \textit{continued training (CT)} using only original or augmented data on LibriSpeech \texttt{test-clean} and \texttt{test-other} data sets with and without shallow fusion (SF).}
\label{tab:asr_librispeech_noorig}
\vspace{-4mm}
\end{table}
}

% ----------------------------------------------------------
%\vspace{-3mm}
\section{Related Work}
\label{sec:related}

Pseudo-labeling \cite{xu2020iterative,chenWW20,zhang2020,xu2021self} is an effective technique to use external models to generate new source-target training pairs from speech sources without transcriptions or target texts without audio. Examples are noisy student training \cite{park2020improved}, consistency training \cite{TjandraS017,HayashiWZTHAT18,hori2019cycle} and TTS-generated data \cite{RosenbergZRJMWW19,WangRcZRWM20,chen2021semi}. Possible disadvantages of pseudo-labeling are its dependency on the quality of the data, and cost of integrating external models or tools, which is not necessary in our approach.

Other techniques generate new labeled data by assembling information solely from the existing training data. For example, MixSpeech \cite{meng2021mixspeech} creates a new audio spectrogram by linearly interpolating two spectrograms. Our method creates new data instances by concatenation in the temporal dimension. This is similar to segmenting audio-target sequences into smaller paired units in the temporal dimension, %. This techniques which has been shown to enhance ASR in low-resource and code-switching settings
for ASR \cite{nguyen2020improving,ye2022improving,lam2021fly} and speech-translation \cite{lam2022sample}. DA by segmentation requires an acoustic aligner, whereas our method does not rely on any external information. % when generating new training instances.

Similar concatenation-based techniques have been applied for special purposes, e.g., random audio concatenation in speech-to-speech translation \cite{jia2022translatotron}, or generating longer inputs for document-level neural machine translation (NMT) \cite{nguyen2021data}. Our work focuses on speech-to-text with the purpose of improving pre-trained models via continued training.

%TK: Better to spend the space in Results
%\begin{itemize}
%    \item \cite{wang2022non}: DA for ASR using voice conversion
%    \item \cite{narayanan2019recognizing}: Memory perturbation in LSTM  
%\end{itemize}

\begin{table*}[ht]
\centering
%\resizebox{.48\textwidth}{!}{%
%\resizebox{\textwidth}{!}{%
\begin{tabular}{lccccc}
\hline
\textbf{Model} & test (En) & test (De) & test (Ca) & test (Fr) & test (Es) \\
%\textbf{Model} & \texttt{test (En)} & \texttt{test (De)} & \texttt{test (Ca)} & \texttt{test (Fr)} & \texttt{test (Es)} \\
\hline
Pre-trained & $19.76$ & $20.47$ & $13.64$ & $15.41$ & $14.66$ \\
\hline
%CT orig $\cup$ CatSelf & $20.75$ & $26.06$ & $14.18$ & $16.05$ & $15.21$ \\
CT orig $\cup$ CatSpeaker & $19.67$ {\footnotesize $\pm 0.00$} & $19.71$ {\footnotesize $\pm 0.02$} & $12.79$ {\footnotesize $\pm 0.18$} & $14.98$ {\footnotesize $\pm 0.00$} & $14.05$ {\footnotesize $\pm 0.04$} \\
CT orig $\cup$ CatRandom & $19.63$ {\footnotesize $\pm 0.13$} & $19.55$ {\footnotesize $\pm 0.04$} & $12.89$ {\footnotesize $\pm 0.02$} & $15.04$ {\footnotesize $\pm 0.07$} & $14.13$ {\footnotesize $\pm 0.05$} \\
\hline
\end{tabular}
%s}
\caption{Word Error Rate of \textit{pre-trained} and \textit{continued training (CT)} ASR models trained on CoVoST-2 English (En), German (De), Catalan (Ca), French (Fr), and Spanish (Es) languages. The ``$\pm$'' values indicate standard deviation over 3 runs.
}
\label{tab:asr_covost2}
\vspace{-3mm}
\end{table*}

\begin{table}[ht]
\centering
\resizebox{.48\textwidth}{!}{%
%\resizebox{\textwidth}{!}{%
\begin{tabular}{lccccc}
\hline
%\textbf{Model} & \texttt{test (En)} & \texttt{test (De)} & \texttt{test (Ca)} & \texttt{test (Fr)} & \texttt{test (Es)} \\
\textbf{Model} & test (En) & test (De) & test (Ca) & test (Fr) & test (Es) \\
\hline
Pre-trained & $19.76$ & $20.47$ & $13.64$ & $15.41$ & $14.66$ \\
\hline
CT orig & $20.10$ & $20.90$ & $13.98$ & $15.38$ & $15.31$ \\
%CT CatSelf & $117.87$ & $118.20$ & $110.32$ & $114.20$ & $112.15$ \\
CT CatSpeaker & $27.36$ & $22.34$ & $12.94$ & $15.69$ & $15.22$ \\
CT CatRandom & $25.46$ & $21.74$ & $13.68$ & $15.69$ & $15.54$ \\
\hline
\end{tabular}
}
\caption{Ablation experiment: Word Error Rate \textit{continued training (CT)} using only original or augmented data on CoVoST-2 English (En), German (De), Catalan (Ca), French (Fr), and Spanish (Es) languages. }
\label{tab:asr_covost2_noorig}
\vspace{-5mm}
\end{table}

\vspace{-2mm}
\section{Method}
\vspace{-2mm}

%Our data augmentation strategy is to concatenate the selected training instances in the temporal dimension, i.e., source-to-source and target-to-target concatenations. There is no special token separating instances in the concatenated instance so the pre-trained model requires no modification in their vocabulary. We explore three simple concatenation strategies: 1) CatSelf, 2) CatSpeaker and 3) CatRandom. CatSelf requires no additional information for augmentation. It resembles ``new'' training data by repeating itself along the temporal dimension. CatSpeaker uses speaker information. Its augmented data contains longer audio spoken by the same person but possibly with dis-coherent content. Similar to CatSelf, CatRandom requires no additional information for augmentation. Different from CatSpeaker, the audio side of its augmented data can be spoken by different people.

Our DA strategy is to concatenate selected training instances in the temporal dimension, i.e., source-source and target-target concatenations. 
%There is no special token separating the original instances in the concatenated instance so there is no need to modify the vocabulary when using a pre-trained off-the-shelf model. 
As there is no special separating token introduced by our method, we can make use of pre-trained off-the-shelf models. 
We evaluate two simple concatenation strategies: %(1) CatSelf, (2) CatSpeaker and (3) CatRandom. 
(1) CatSpeaker makes use of speaker information and generates longer audio-text pairs spoken by the same person. 
(2) CatRandom generates new training instances by randomly concatenating audio-text pairs, spoken by different persons. %Thus, it requires no additional information similar to CatSelf.
We also tried out a third concatenation strategy that generates new training instances by repeating the original instance along the temporal dimension. This method, however, resulted in spurious repetitions in the output, consistently producing worse results than the pre-trained models. %, thus requiring no additional information for augmentation. 

%We create augmented data on-the-fly. At the beginning of each epoch, we generate a new batch iterator and allow concatenation be done over the entire training data. Then, we combine the original training data (orig ) and the augmented data, apply filtering on it before generating the batches. In contrast to creating new instances by concatenating with other instances in the same batch, this helps to increase the diversity of the augmented data, and reduces the out-of-memory error caused by the lengthy augmented samples. Finally, this combined data would be used for continued training \textit{CT} the pre-trained model or training a model from scratch \textit{FS}

Our approach applies data augmentation \textit{on-the-fly}. At the beginning of each epoch, we 
%generate a new batch iterator and 
allow concatenations over the entire training data. Then, we combine the original training data and the augmented data, and apply length filtering before generating the training batches. 
By allowing concatenations over the entire training data instead of over only the current batch, we increase diversity 
of the augmented data. %Furthermore, filtering at the beginning of each epoch avoids out-of-memory errors caused by lengthy augmented samples. 
This concatenated data is then used for \textit{continued training} of pre-trained models or training new models \textit{from scratch}. 

% ----------------------------------------------------------
\vspace{-2mm}
\section{Experimental Setup}
\vspace{-2mm}
% ----------------------------------------------------------
\subsection{Datasets and preprocessing}

For the ASR tasks, we evaluate our method on LibriSpeech \cite{panayotovETAL:15} and the CoVoST-2 \cite{WangETAL:20} ASR dataset. For LibriSpeech, we combine the transcriptions in train960h and the extra 800M-word monolingual text data to train the LM. For CoVoST-2 ASR, we test on five languages: English (En), German (De), Catalan (Ca), French (Fr) and Spanish (Es). For the automatic speech translation (AST) tasks, we evaluate our method on CoVoST-2 and MuST-C for En-De. On both dataset, we use their own transcription-translation training data to train NMT models for knowledge distillation \cite{inagumaETAL:21}. 
%Table \ref{tab:data_statistics} shows the statistics of the above corpora used in the experiments. 

For all speech inputs, we extracted 80-dimensional log Mel-filterbank with 25ms FFT windows and 10ms frame shift. We filter instances with more than 3k frames. For transcriptions in LibriSpeech, we use the vocabulary file of 10k subword units from the \textsc{FairSeq} GitHub repository.\footnote{\url{https://github.com/facebookresearch/fairseq}} For CoVoST-2 ASR tasks, we lowercased transcriptions and removed punctuation. For each language, we use 5k subword units. For translation tasks, we do not apply preprocessing on the translation data. For NMT and AST, the size of subword units are 5k and 8k for CoVoST-2 and MuST-C, respectively. 
%For NMT, we create an universal vocabulary for both En and De whereas we use De only in AST. 
All sub-word units are built using SentencePiece \cite{kudo2018sentencepiece}.

%\begin{table}[t]
%\centering
%\resizebox{.48\textwidth}{!}{%
%\begin{tabular}{lccccc}
%\hline
%\textbf{Data} & Lang. & Tokens (unique) & Sentences & Tokens/Sentence & Speakers \\
%\hline
%LibriSpeech & En & $9.4$M ($89$k) & $281$k & $33.4$ {\footnotesize $\pm 11.9$} & $2338$ \\
%\hline
%CoVoST-2 & En & $2.8$M ($126$k) & $289$k & $9.80$ {\footnotesize $\pm 3.0$} & $10$k \\
%CoVoST-2 & De & $1.1$M ($88$k) & $128$k & $8.69$ {\footnotesize $\pm 3.0$} & $1070$ \\
%CoVoST-2 & Ca & $939$k ($45$k) & $95.8$k & $9.80$ {\footnotesize $\pm 4.6$} & $557$ \\
%CoVoST-2 & Fr & $1.8$M ($104$k) & $207$k & $8.79$ {\footnotesize $\pm 3.4$} & $1754$ \\
%CoVoST-2 & Es & $741$k ($50$k) & $79$k & $9.38$ {\footnotesize $\pm 3.14$} & $1197$ \\ 
%\hline
%\end{tabular}
%}
%\caption{Statistics of the training data used in the experiments}
%\label{tab:data_statistics}
%\end{table}

\subsection{Model Architectures}
We use \textsc{FairSeq} \cite{ott2019fairseq,wang2020fairseqs2t} for our implementation. For LibriSpeech, we used a pre-trained Transformer-based ASR model labeled \textit{s\_transformer\_l} downloaded from the \textsc{FairSeq} GitHub repository mentioned above. 
%This ASR model has about 270M parameters. 
%It has two 1D convolution layers for down-sampling the input sequence-length by a factor of 4, 12 transformer-encoder layers and 6 transformer-decoder layers. The attention dimension and the feed-forward network (FFN) dimension is 1024 and 4096, respectively. 
%Further details can be found in the documentation.
For shallow fusion, we use a Transformer-based LM of about 24M parameters. It has 6 layers with attention dimension of 512 and with FFN dimension of 2048.
For CoVoST-2 ASR \& AST and MuST-C AST, we used a Conformer architecture \cite{gulati2020conformer}, labeled as \textit{s2t\_conformer}, of about 45M parameters. %It also has two 1D convolution layers for down-sampling the inputs by a factor of 4. It has 12 conformer-encoder layers and 6 decoder layers. The kernel size of depth-wise convolution, the attention dimension, and the FFN dimension, is 31, 256 and 2048, respectively. We use attention type ``attn-type=espnet`` and absolute positional encoding. 
We follow the default configuration, with the exception of using 12 encoder layers and using attention type ``attn-type=espnet``. For NMT, we use a transformer of encoder-decoder-layers of size 3 and 6 for CoVoST-2 and MuST-C, respectively. Dimensions of attention and FFN-layer are 256 and 2048, respectively.

%\vspace{-8mm}

\subsection{Training and Inference}

We use Adam optimizer \cite{kingma2014adam} with inverse square root learning rate schedule for all experiments. % The peak learning rate (lr) and the number of warm-up steps are adjusted per experiment. In continued training, we reset the optimizer with a lr of 2e-3 and 1k steps in warm-up. 
For all experiments, we use a peak learning rate (lr) of 2e-3, with the exception of LM and NMT training where we use a lr of 5e-4 and of 1e-3, respectively. For pre-training and training from scratch, we adjust the warm-up steps for different settings. For continued training, we reset the optimizer with 1k warm-up. % steps. 
All speech-to-text experiments use a batch size of 40k$\times$8 frames for training except for MuST-C, where we use 40k$\times$2 and 25k$\times$8 for ASR and AST, respectively. SpecAugment \cite{ParkCZCZCL19} is applied with a frequency mask of 27 and a time mask parameter of 100, with 2 masks along their respective dimension.

For LibriSpeech, we examined our strategies by training the pre-trained ASR model for 50k steps with validation step of 2k. The LM is trained for 200k steps with a batch size of 16k$\times$2 tokens with 4k warm-up steps. For both ASR and LM, decoder-input and output embedding are shared. % during training. %during half precision (fp16) training. 

For CoVoST-2 ASR, the pre-trained ASR models and the FC cases are trained for 30k steps, validated by every 500 steps. The exception is English which has more data. We thus train it for 60k steps with a validation step of 1k. 
All above models use 10k warm-up steps. For continued training, the En-ASR is trained for 20k steps, validated every 1k steps. De-ASR and Fr-ASR are trained for 10k steps whereas Ca-ASR and Es-ASR are trained for 8k steps. These four language pairs are validated every 500 steps. %For AST, we initialize the encoder with a pre-trained En-ASR. The AST is then trained for 50k steps with 10k warm-up and validated every 1k steps.

%For MuST-C, we first use the pre-trained En-ASR\footnote{For both CoVoST-2 and MuST-C, the En-ASR models used in initialisation are trained on the original data only. In addition, the ASR models are obtained by averaging their 5 best checkpoints on their validation losses.} %with a batch size of 40k$\times$2 frames 
%for 100k steps. The AST is then trained for 100k steps for each strategy %with a batch size of 25k$\times$8 frames. 
For MuST-C, both ASR\footnote{For both CoVoST-2 and MuST-C, the En-ASR models used in initialisation are trained on the original data only. In addition, the ASR models are obtained by averaging their 5 best checkpoints on their validation losses.} and AST use 100k steps in training with 25k in warm-up and every 2k steps in validation. The NMT\footnote{CoVoST-2 NMT is similar except of a batch size of 16k tokens.} is trained for 100 epochs with 8k warm-up steps, validated every epoch, and with a batch size of 100 sentences. For CoVoST-2 AST, we initialize the encoder with a pre-trained En-ASR. The AST is then trained for 50k steps with 10k warm-up and validated every 1k steps.

%For inference, 
We use beam search of size 5 during inference. In shallow fusion, we use the last checkpoint with an interpolation weight of 0.3. For pre-training and training from scratch, we average the best 5 checkpoints by validation loss. For continued training, we average the \textit{last} 5 checkpoints per validation step to prevent the averaging over pre-training checkpoints. For AST, we again average over the best 5 checkpoints. 

% ----------------------------------------------------------
\section{Experimental Results}
% ----------------------------------------------------------

% ----------------------------------------------------------
\subsection{LibriSpeech}
% ----------------------------------------------------------

%Table \ref{tab:asr_librispeech} shows the Word-Error-Rate (WER) of CT \todo{consistent notation: texttt or not?}for each of the proposed concatenation strategies. CT on the joint data created by CatSelf shows worse performance in all settings. The WER degradation ranges from 0.46 to 1.11. 
%Both CatSpeaker and CatRandom show substantial improvements over the baseline system, with CatSpeaker performing slightly better than CatRandom. CatSpeaker has a maximum reduction of 0.47 WER and of 0.64 WER on the \texttt{test-clean} and \texttt{test-other} splits, respectively. 
Table \ref{tab:asr_librispeech} lists Word-Error-Rate (WER) of the continued training experiments for each of the proposed concatenation strategies. 
%CatSelf shows the worst performance in all settings and deteriorates even over the baseline model, resulting in WER degradation from 0.46 to 1.11. 
Both CatSpeaker and CatRandom show significant improvements over the baseline system, with CatSpeaker performing slightly better than CatRandom throughout the experiments. 
We conjecture that speaker information is useful for ASR in the audiobooks domain, but the effect is very limited. 
Compared to the baseline that is trained on the original data only, CatSpeaker shows a reduction of 0.47 WER (14.2\% relative) and of 0.64 WER (8.5\% relative) on the \texttt{test-clean} and \texttt{test-other} splits, respectively. 
Further improvements can be achieved by using shallow fusion in decoding, resulting in 2.55 WER on \texttt{test-clean} (18.5\% relative reduction) and 6.27 WER on \texttt{test-other} (7.9\% relative reduction). 
All improvements over the pre-trained model are significant with $p <$ 0.005 according to an approximate randomization test \cite{riezlermaxwell:05}.

%Table \ref{tab:asr_librispeech_noorig} shows an ablation study of contributions of each data set under the CT strategy. CT orig refers to continued training on the original training data set by the same number of updates as the augmented one. In general, it shows worse or slight improvement over the baseline. The biggest improvement is on \texttt{test-other} which has an improvement of 0.13 WER. Continued training on the augmented dataset CatSelf only shows substantially worse performance over the baseline. The system performance is degraded to the level 41 WER or worse in all settings. We find that the system has a tendency to repeat the hypothesis itself. Interestingly, both continued training on CatSpeaker and CatRandom yield similar improvements as the joint data setting, which itself yields a further improvement 0.15 WER on \texttt{test-other}.

Table \ref{tab:asr_librispeech_noorig} shows an ablation study where training is continued using only augmented data without adding the original data. ``CT orig''  refers to continued training on the original training data set by the same number of updates as the augmented one. Here, we observed only minimal to no improvements.
%Continued training on the CatSelf data only shows largely worse performance compared to the baseline. A detailed inspection of the generated transcriptions reveals the underlying problem: The Transformer-based system tends to frequent repetitions in its output, a property that is introduced by this type of augmented data. 
Continued training on both CatSpeaker and CatRandom yields similar improvements with and without the inclusion of the original data. 

%In general, it shows worse or slight improvement over the baseline. The biggest improvement is on \texttt{test-other} which has an improvement of 0.13 WER. Continued training on the augmented dataset CatSelf only shows substantially worse performance over the baseline. The system performance is degraded to the level 41 WER or worse in all settings. We find that the system has a tendency to repeat the hypothesis itself. 

%Interestingly, both continued training on CatSpeaker and CatRandom yield similar improvements as the joint data setting, which itself yields a further improvement 0.15 WER on \texttt{test-other}.

\vspace{-1mm}
% ----------------------------------------------------------
\subsection{CoVoST-2}
% ----------------------------------------------------------
\vspace{-1mm}

%Table \ref{tab:asr_covost2} shows the WER of the concatenation strategies using CT on 5 languages of the CoVoST-2 dataset. Similar to LibriSpeech, CT on the joint data set of orig and CatSelf yields worse WER than the pre-trained models for all the 5 languages. The degradation of WER ranges from 0.54 points to 5.59 points with De having the largest degradation. Both CatSpeaker and CatRandom yield similar WER improvements for each language pair. In all languages, both CatSpeaker and CatRandom shows improvements over the pre-trained model with a maximum WER improvement of 0.92 points for De. However, the improvement on English is rather marginal, i.e., 0.13 WER using CatRandom in the best case. We attribute this to the larger amount of the training data but simpler sentence structure. 

\begin{table*}[t]
\centering
%\resizebox{.48\textwidth}{!}{%
%\resizebox{\textwidth}{!}{%
\begin{tabular}{lccccc}
\hline
%\textbf{Model} & \texttt{test (En)} & \texttt{test (De)} & \texttt{test (Ca)} & \texttt{test (Fr)} & \texttt{test (Es)} \\
\textbf{Model} & test (En) & test (De) & test (Ca) & test (Fr) & test (Es) \\
\hline
Pre-trained & $19.64$ {\footnotesize $\pm 0.09$} & $20.40$ {\footnotesize $\pm 0.07$} & $13.58$ {\footnotesize $\pm 0.09$} & $15.35$ {\footnotesize $\pm 0.05$} & $14.79$ {\footnotesize $\pm 0.18$} \\
\hline
%FS orig $\cup$ CatSelf & $21.59$ & $21.47$ & $13.67$ & $16.47$ & $15.53$ \\
FS orig $\cup$ CatSpeaker & $19.65$ {\footnotesize $\pm 0.04$} & $19.50$ {\footnotesize $\pm 0.05$} & $12.09$ {\footnotesize $\pm 0.19$} & $14.87$ {\footnotesize $\pm 0.11$} & $14.03$ {\footnotesize $\pm 0.11$} \\
FS orig $\cup$ CatRandom & $19.44$ {\footnotesize $\pm 0.17$} & $19.22$ {\footnotesize $\pm 0.00$} & $11.96$ {\footnotesize $\pm 0.12$} & $14.94$ {\footnotesize $\pm 0.08$} & $14.14$ {\footnotesize $\pm 0.06$} \\
\hline
\end{tabular}
%}
\caption{Word Error Rate of different ASR systems trained \textit{from scratch (FS)} on the ASR part of CoVoST-2 English (En), German (De), Catalan (Ca), French (Fr) and Spanish (Es) languages. 
The ``$\pm$'' values are standard deviations over 3 runs.
%The checkpoints used is the average of the best 5 checkpoints against the validation loss
}
\label{tab:asr_covost2_scratch}
\vspace{-6mm}
\end{table*}

Table \ref{tab:asr_covost2} lists the WER of our concatenation strategies with continued training on 5 languages of the CoVoST-2 dataset. 
%We see similar results to the LibriSpeech experiments, where CatSelf results in worse WER than the pre-trained models on all 5 languages. The degradation in WER ranges from 0.54 points for Catalan to 5.59 points for German, where the pre-trained systems is best for  Catalan and worst for German. 
%
%Similar to LibriSpeech, CT on the joint data set of orig and CatSelf yields worse WER than the pre-trained models for all the 5 languages. The degradation of WER ranges from 0.54 points to 5.59 points with De having the largest degradation. Both CatSpeaker and CatRandom yield similar WER improvements for each language pair. In all languages, both CatSpeaker and CatRandom shows improvements over the pre-trained model with a maximum WER improvement of 0.92 points for De. However, the improvement on English is rather marginal, i.e., 0.13 WER using CatRandom in the best case. We attribute this to the larger amount of the training data but simpler sentence structure. 
%
The CatSpeaker and CatRandom strategies yield similar WER improvements for each language. However, there is no consistent trend that might indicate if speaker information is useful or not. 
Throughout all languages, both CatSpeaker and CatRandom shows improvements over the pre-trained model with the largest WER improvement of 0.92 points (4.4\% relative) for German, and the largest relative improvement in WER of 6.2\% (0.75 points absolute) for Catalan. At the same time, the improvement on English is rather marginal even in the best case, i.e., 0.13 WER (0.7\% relative) for CatRandom. 
We attribute this to the larger amount of the English training data compared to the other languages. 
The fact that this observation differs from the ASR improvements on LibriSpeech can be explained by the much simpler sentence complexity of the CoVoST-2 data.
All improvements over the pre-trained model are significant with $p <$ 0.002 except for English \cite{riezlermaxwell:05}.

% ----------------------------------------------------------
%\subsubsection{Ablation experiment}
% ----------------------------------------------------------
%\vspace{-1mm}

In Table \ref{tab:asr_covost2_noorig} we repeat our ablation experiment to evaluate the contribution of the augmented data only. 
%Similar to LibriSpeech, continued training using CatSelf data shows the worst performance compared to the baseline. An analysis of the transcriptions again reveals that the models tend to spurious repetitions in the output.
In all cases except French, continued training using the original data also slightly degrades the model compared to the baseline. This is likely due to overfitting, as the pre-trained models use checkpoint-averaging to improve generalization, which is then reduced by continued training. 
Unlike the previous results on LibriSpeech, training on augmented data created by CatSpeaker and CatRandom mostly show worse performance over the pre-trained model. A slight improvement can be observed only for Catalan using the CatSpeaker data. 
%We conjecture that the inclusion of the original data is vital for continued training on this dataset. 
%Training only on augmented data mostly shows worse performance over the pre-trained model. 
%A length-dependent analysis of the errors revealed that spurious textual repetitions occur especially for short sentences. 
%This is a well-known problem when Transformer-based models are not trained on enough short examples. 
%A length-dependent analysis revealed that the errors occur especially on shorter examples, which can partly be ascribed to length distribution differences between training and test sets \cite{WanETAL:2022}. Furthermore, our augmented data does not contain any single token examples, but such are present in CoVoST-2 test. %which are frequent in CoVoST-2.
%This problem is effectively tackled by including the original data. 
%e shorter CoVoST-2 test examples are much more susceptible to this effect than LibriSpeech e. % with an average length of 33.4. % tokens.

\begin{figure}[t]
     \centering
     \begin{subfigure}[b]{0.239\textwidth}
         \centering
\includegraphics[angle=-90,width=\textwidth]{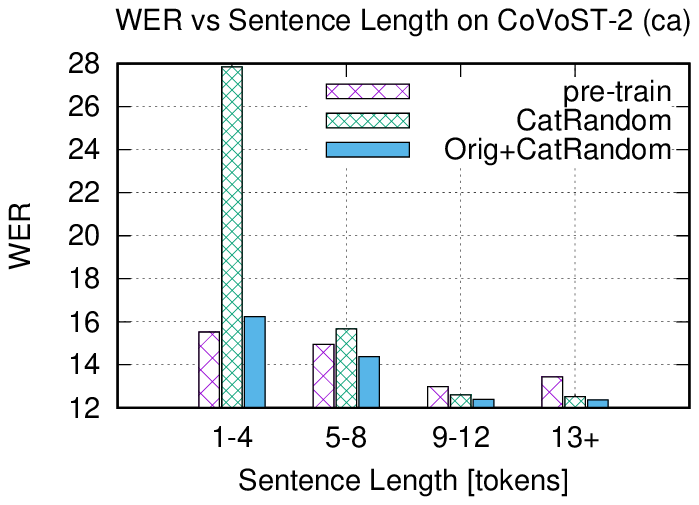}
     \end{subfigure}
     \hfill
     \begin{subfigure}[b]{0.239\textwidth}
         \centering
\includegraphics[angle=-90,width=\textwidth]{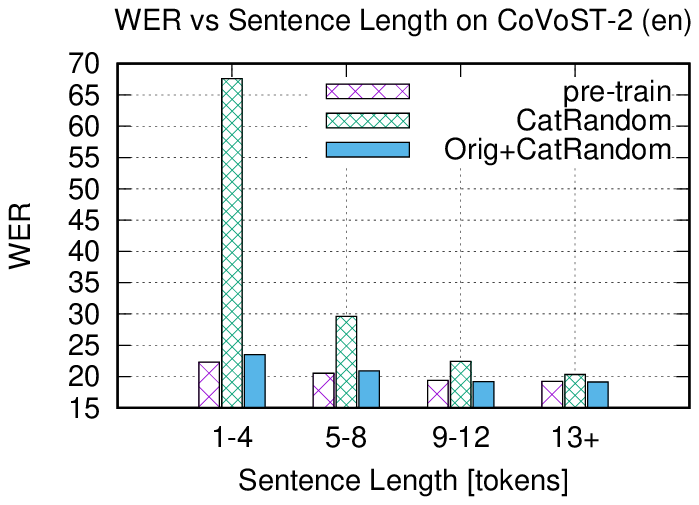}
     \end{subfigure}
        \caption{WER w.r.t. sentence length on CoVoST-2.}
        \label{fig:length_analysis}
\vspace{-6mm}
\end{figure}

% ----------------------------------------------------------
%\subsubsection{Training from scratch}
% ----------------------------------------------------------

Finally, we evaluate our concatenation strategies by training the entire ASR model from scratch for each language. Table \ref{tab:asr_covost2_scratch} lists the results. 
For most cases, the improvements obtained by training from scratch are very close to those by continued training. 
Only for Catalan we observe further WER reduction of 0.86 compared to the continued training. 
Thus, our method also works for training from scratch if such training resources are available. Alternatively, one can use an off-the-shelf model and improve it via continued training with our method consuming much less computing power.
%In all cases except Catalan, the improvement obtained by training from scratch is similar to that of continued training. For Catalan, we observe further WER reduction of 0.86 compared to the continued training. 
%\todo{training from scratch for LibriSpeech?}

\vspace{-2mm}
\subsection{Length Dependent Analysis} 
\vspace{-1mm}

We conducted a deeper analysis of the ablation study in Table \ref{tab:asr_librispeech_noorig} and in Table \ref{tab:asr_covost2_noorig} by evaluating test examples based on their length: training using only augmented data leads to a strong increase in WER for short examples, where spurious repetitions of the textual output is the most noticeable problem. By concatenating examples, the data length distribution is shifted to the longer side and changed particularly at the ends; e.g., examples containing only 1 token are completely absent in the augmented data. Furthermore, CoVoST-2 has about 9\% test examples of 5 tokens or less, whereas LibriSpeech has only 1.5\%. The increase of errors on short examples thus affects the overall WER much more on the CoVoST-2 dataset. 
%Including the original data during training effectively reduces this problem as Figure \ref{fig:length_analysis} illustrates for Catalan (left) and English (right). Here, we plot WER with respect to sentence length for the pre-trained models, and for continued training on CatRandom and on original+CatRandom data. The figures for the other languages look very similar.
In Figure \ref{fig:length_analysis} we plot WER on CoVoST-2 w.r.t. sentence length for the pre-trained models, and for continued training on CatRandom and on original+CatRandom data. Including the original data during training effectively reduces this problem as can be seen from Figure \ref{fig:length_analysis} for Catalan (left) and English (right), and we found a similar behavior for the other three languages.

\vspace{-2mm}
% ----------------------------------------------------------
\subsection{AST (En-De): MuST-C and CoVoST-2} 
% ----------------------------------------------------------
\vspace{-1mm}

We also evaluate our proposed DA strategies on two En-De speech-to-text translation tasks. Table \ref{tab:ast_mustc+covost2_chrf2} lists the chrF2 \cite{popovic2015chrf} scores of systems trained with ``orig'' (original data plus translations generated by knowledge distillation) and trained with combined data created by CatSpeaker or by CatRandom. %Both concatenation strategies achieve significant improvement with $p$-value $<$ 0.0002 on MuST-C tst-COMMON and with $p$-value $<$ 0.0001 on CoVoST-2 test set.
Both concatenation strategies achieve significant improvements with $p <$ 0.00025 both on MuST-C tst-COMMON and on CoVoST-2 test sets using the approximate randomization test implementation of \textsc{sacreBLEU}\footnote{\scriptsize{}nrefs:1|ar:10000|seed:12345|case:mixed|eff:yes|nc:6|nw:0|space:no|version:2.0.0} \cite{post:18}.

%\vspace{-1mm}
\begin{table}[h!]
\centering
%\resizebox{.48\textwidth}{!}{%
\resizebox{.48\textwidth}{!}{%
\begin{tabular}{lcc}
\hline
\textbf{Model} & MuST-C tst-COMMON & CoVoST-2 test \\
\hline
orig & $52.8$ {\footnotesize $\pm 0.0$} & $47.65$ {\footnotesize $\pm 0.05$} \\
orig $\cup$ CatSpeaker & $53.55$ {\footnotesize $\pm 0.05$} & 48.55 {\footnotesize $\pm 0.05$}  \\
orig $\cup$ CatRandom & $53.55$ {\footnotesize $\pm 0.05$} & 48.45 {\footnotesize $\pm 0.05$} \\
\hline
\end{tabular}
}
\caption{chrF2 on MuST-C AST and CoVoST-2 AST (En-De). The ``$\pm$'' values indicate standard deviations over 2 runs.}
\label{tab:ast_mustc+covost2_chrf2}
\vspace{-2mm}
\end{table}

%\begin{table}[ht]
%\centering
%\resizebox{.48\textwidth}{!}{%
%\resizebox{.48\textwidth}{!}{%
%\begin{tabular}{lcc}
%\hline
%\textbf{Model} & \texttt{MuST-C tst-COMMON} & \texttt{CoVoST-2 test} \\
%\hline
%Orig & 25.79 ± 0.04 & 21.43 ± 0.00 \\
%Orig $\cup$ CatSpeaker & 26.16 ± 0.02 & 21.94 ± 0.13  \\
%Orig $\cup$ CatRandom & 26.01 ± 0.04 & 21.92 ± 0.01 \\
%\hline
%\end{tabular}
%}
%\caption{BLEU on MuST-C AST and CoVoST-2 AST (En-De) (2 runs)}
%\label{tab:ast_mustc+covost2}
%\end{table}

The results show that our simple method is also applicable to AST where the speech-text alignments are not parallel.

\vspace{-1mm}
% ----------------------------------------------------------
\section{Conclusion}
% ----------------------------------------------------------
\vspace{-1mm}
We propose and evaluate temporal-concatenation as a data augmentation method for improving Transformer and Conformer based speech-to-text models. The method can be applied to improve pre-trained models without requiring extra information or external tools. We evaluate three concatenation strategies for ASR on LibriSpeech and CoVoST-2 data and found that concatenation by random and concatenation by speaker perform similarly and bring significant improvements.
Finally, we evaluate our method for AST on Must-C and CoVoST-2 and also observed significant improvements.
In the future, we would like to extend our method to other architectures and tasks.

\vspace{-1mm}
% ----------------------------------------------------------
\section{Acknowledgements}
% ----------------------------------------------------------
\vspace{-1mm}
This research was supported in part by the German
research foundation DFG under grant RI-2221/4-1.
% -------------------------------------------------------------------------
\bibliographystyle{IEEEbib}
{
\footnotesize
\bibliography{references}
}
\end{document}